\documentclass[journal]{IEEEtran}
\usepackage{cite}
\usepackage[cmex10]{amsmath}
\usepackage{breqn}
\usepackage{algorithmic}
\usepackage{array}
\usepackage{fixltx2e}
\usepackage{url}
\usepackage{hyperref}
\usepackage{graphics}
\usepackage{graphicx}
\usepackage{amssymb}
\usepackage{multirow}
\usepackage{tabularx,multicol,multirow}
\usepackage[dvipsnames,table]{xcolor}

\newcolumntype{C}{>{\centering\arraybackslash}X}

\begin{document}

\title{BASS Net: Band-Adaptive Spectral-Spatial Feature Learning Neural Network for Hyperspectral Image Classification}

\author{Anirban~Santara*,~\IEEEmembership{}
        Kaustubh~Mani*,~\IEEEmembership{}
        Pranoot~Hatwar,~\IEEEmembership{}
        Ankit~Singh,~\IEEEmembership{}
        Ankur~Garg,~\IEEEmembership{}
        Kirti~Padia~\IEEEmembership{}
        and~Pabitra~Mitra~\IEEEmembership{}
\thanks{* Denotes equal contribution.}
\thanks{
A. Santara and P. Mitra are with the Department
of Computer Science and Engineering, Indian Institute of Technology, Kharagpur,
WB, 721302 India (e-mail: anirban\_santara@iitkgp.ac.in, pabitra@cse.iitkgp.ernet.in)}
\thanks{P. Hatwar and A. Singh are with the Department
of Electrical Engineering, Indian Institute of Technology, Kharagpur,
WB, 721302 India (e-mail: pphatwar1995@gmail.com, ankit.vngr@gmail.com)}
\thanks{K. Mani is with the Department of Geology and Geophysics, Indian Institute of Technology, Kharagpur,
WB, 721302 India (e-mail: kaustubh3095@gmail.com)}
\thanks{A. Garg and K. Padia are with Space Applications Centre, Indian Space Research Organization (ISRO) Ahmedabad,
GJ, 380015 India (e-mail: agarg@sac.isro.gov.in, kirtipadia@sac.isro.gov.in)}
}

\markboth{Submitted to IEEE Transactions on Geoscience and Remote Sensing}%
{Santara \MakeLowercase{\textit{et al.}}: BASS Net: Band-Adaptive Spectral-Spatial Feature Learning Neural Network for Hyperspectral Image Classification}

\maketitle
\begin{abstract}
Deep learning based landcover classification algorithms have recently been proposed in literature. In hyperspectral images (HSI) they face the challenges of large dimensionality, spatial variability of spectral signatures and scarcity of labeled data. In this article we propose an end-to-end deep learning architecture that extracts band specific spectral-spatial features and performs landcover classification. The architecture has fewer independent connection weights and thus requires lesser number of training data. The method is found to outperform the highest reported accuracies on popular hyperspectral image data sets.
\end{abstract}


\IEEEpeerreviewmaketitle

\section{Introduction}
\label{sec:intro}
Hyperspectral imaging \cite{Landgrebe:2002,Richards:2013} collects rich spectral information from a large number of densely-spaced contiguous frequency bands. It produces three dimensional $(x, y, \lambda)$ data volumes, where $x,y$ represent spatial dimensions and $\lambda$ represents spectral dimension. \\

Hyperspectral image classification \cite{Tuia:2013} is the task of assigning a class label to every pixel. This paper studies land-cover classification in hyperspectral images where the task is to predict the type of land-cover present in the location of each pixel. There are several challenges associated with hyperspectral data the most critical of which are as follows \cite{Valls:2005}.
\begin{enumerate}
\item Curse of dimensionality resulting from large number of spectral dimensions.
\item Scarcity of labelled training examples. 
\item Large spatial variability of spectral signature.
\end{enumerate}
\vspace{4.5mm}

\indent Several approaches have been followed in literature for HSI classification. The simplest of them are based on $k$-nearest neighbors ($k$-NN). In these methods, given a test sample, Eucledian distance in the input space or a transformed space is used to find the $k$ nearest training examples and a class is assigned on the basis of them. In \cite{Blanzieri:2008} and \cite{Li:2015} some modified versions of the $k$-NN algorithm have been proposed for HSI classification. Support Vector Machine (SVM) classifier is a maximum margin linear classifier \cite{Burges:1998}. Melghani et al.\cite{Melgani:2004} introduced SVM Classifier for HSI classification. SVM based methods, in general, follow a two step approach.
\begin{enumerate}
\item Dimensionality reduction in order to address the problems of high spectral dimensionality and scarcity of labeled training examples. Some of the methods followed for dimensionality reduction are subspace projection \cite{Gao:2015}, random feature selection \cite{Waske:2010} and Kernel Local Fisher Discriminant Analysis \cite{Li:2011}.
\item Classification in the reduced dimensional space using SVM \cite{Li:2012,Melgani:2004,Gao:2015}.
\end{enumerate}
Li et al. \cite{Li:2012} propose local Fisher's Discriminant Analysis for dimensionality reduction and Gaussian Mixture Model (GMM) for classification. Mianji et al. \cite{Mianji:2011} propose Gaussian Non-linear Discriminant Analysis for dimensionality reduction and Relevance Vector Machine for classification. Samat et al. \cite{Samat:2014} introduced Extreme Learning Machine (ELM) for HSI classification. ELM \cite{Huang:2006} is a two layer artificial neural network in which the input to hidden weights are randomly chosen and the hidden to output weights are learned by minimizing a least squares objective function. In \cite{Li:2015} LBP is used to extract texture based local descriptors which are combined with global descriptors like Gabor and spectral features and fed into an ELM for classification. Lu et al. \cite{Lu:2016} proposed a set-to-set distance based method for HSI classification.\\

Recently deep neural networks \cite{LeCun:2015,Goodfellow:2016} have been employed for landcover classification in HSI. Deep learning methods for HSI classification \cite{Zhang:2016} focus on spectral-spatial context modeling in order to address the problem of spatial variability of spectral signatures. They fall into two broad categories. The first category \cite{Chen:2014,Ma:2015,Chen:2015,Zhang:2015,Ma:2016} follows a two-step procedure.
\begin{enumerate}
\item Dimensionality reduction and spectral-spatial feature learning using Autoencoder. Autoencoder \cite{Hinton:2006} is an artificial neural network architecture that learns to reconstruct the input vector at the output with minimum distortion after passing through a bottleneck. The vector of activations in the bottleneck is a reduced dimensional representation of the input vector that often encodes useful semantic information.
\item Classification using multi-class logistic regression.
\end{enumerate}
The second category of methods use Convolutional Neural Networks (CNN) \cite{LeCun:1998,Goodfellow:2016} for feature learning and classification in an end-to-end fashion. CNN uses extensive parameter-sharing to tackle the curse of dimensionality. Hu et al. \cite{Hu:2015} introduced CNN for HSI classification. The proposed architecture is designed to learn abstract spectral signatures in a hierarchical fashion but does not take into account spatial context. In \cite{Zhao:2016} compressed spectral features from a local discriminant embedding method are concatenated with spatial features from a CNN and fed into a multi-class classifier. Yu et al.\cite{Yu:2016} and Chen et al.\cite{Chen:2016} propose end-to-end CNN architectures for spectral-spatial feature learning and classification. In \cite{Li:2016}  the idea of classifying pixel-pair features using CNN is introduced to compensate for data scarcity. Also a voting strategy is proposed for test time to provide robustness in heterogeneous regions. \\

In this paper we present a deep neural network architecture that learns band-specific spectral-spatial features and gives state-of-the-art performance without any kind of data-set augmentation or input pre-processing. The architecture consists of three cascaded blocks. Block 1 takes a $p\times p\times N_c$ input volume ($N_c=$ number of spectral channels) and performs a preliminary feature transformation on the spectral axis. It splits the spectral channels into bands and feeds to Block 2 where parallel neural networks are used to extract low and mid-level spectral-spatial features. The outputs of the parallel networks are fused by concatenation and fed into Block 3 which summarizes them to form a high-level representation of the input. This is eventually classified by logistic regression. Extensive use of convolutional layers and weight sharing among the parallel networks of Block 2 keeps the parameter budget and computational complexity low. Band-specific representation learning and fusion via concatenation in Block 2 makes the network discriminative towards spectral locality of low and mid level features. Experiments on benchmark hyperspectral image classification data sets show that the proposed network converges faster and gives superior classification performance than other deep learning based methods in literature. Our source code is publicly available on the Web\footnote{https://github.com/kaustubh0mani/BASS-Net}.\\

The contributions of this paper can be summarized as follows:\\

\begin{enumerate}
\item A novel end-to-end neural network architecture has been proposed that shows state-of-the-art performance on benchmark hyperspectral image classification data sets. The design is aimed at efficient band-specific feature learning keeping the number of parameters low.\\

\item Considerble improvement in training time is observed when compared to other popular deep learning architectures.\\ 

\end{enumerate}

Section \ref{sec:framework} gives a detailed description of the proposed architecture along with the design methodology followed. Experimental results are presented in Section \ref{sec:results}. Comparison with existing methods is also reported. Section \ref{sec:conclusion} concludes the paper with a summary of the proposed method and scope of future work.

\section{Proposed Framework}
\label{sec:framework}

\begin{figure*}
\centering
	\includegraphics[width=0.75\textwidth]{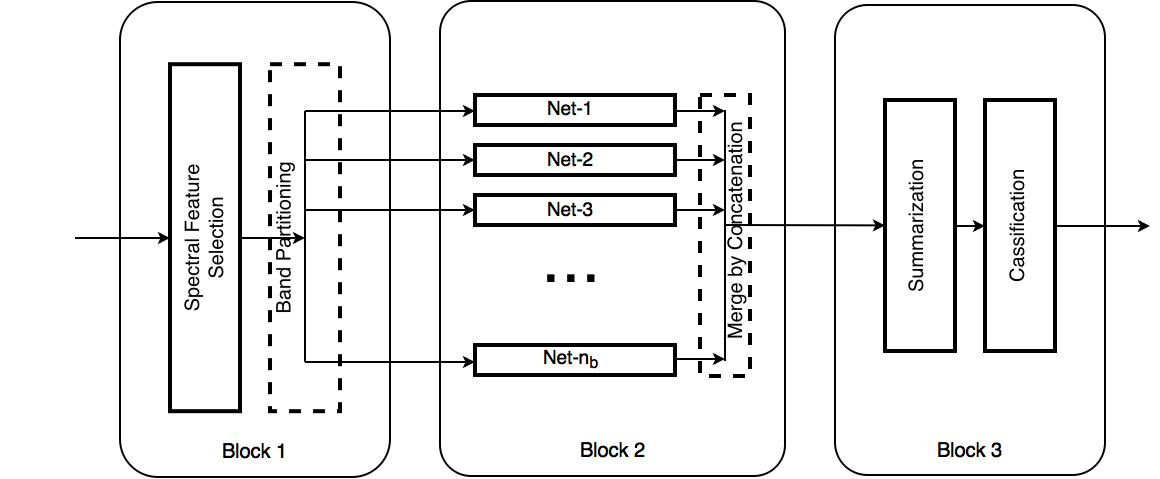}
	\caption{Block diagram of the BASS Net architecture.}
	\label{fig:proposed_arch}
\end{figure*}

The BASS Net architecture, shown in Figure \ref{fig:proposed_arch}, combines spectral and spatial information processing in a systematic way with a focus on efficient use of parameters. The input to the network is a pixel $\mathbf{X_i}$ from the image with its $p\times p$ neigbourhood (for spatial context) in the form of a $p \times p \times N_c$ volume, where $N_c$ is the number of channels in the input image. The output is the predicted class label $\hat{y}_i$ for $\mathbf{X_i}$. The entire network is differentiable end-to-end and can be trained by backpropagation \cite{Rumelhart:1989}. 

\subsection{Overview of architecture}
The architecture is organized as three cascaded blocks.\\

\subsubsection{Block 1: Spectral feature selection and band partitioning}\hfill \\

Block 1 takes the input $p\times p\times N_c$ volume $\mathbf{X_i}$ and performs the following operation.

\begin{equation}
\{B_1, B_2, \dots, B_{n_b}\} = \Psi(\Phi(\mathbf{X_i}), n_b)
\end{equation}

\noindent $\Psi(\cdot,\cdot)$ is a function that takes as input a hyperspectral image volume $\mathbf{X}$ with $N$ spectral channels and an integer $n_b$. It splits $\mathbf{X}$ into $n_b$ non-overlapping adjacent bands $\{B_i\}_{i=1}^{n_b}$ of equal bandwidth $b$, where $b = \frac{N}{n_b}$.
\begin{equation}
\{B_1, B_2, \dots, B_{n_b}\} = \Psi(\mathbf{X}, n_b)
\end{equation}

\noindent $\Phi(\cdot)$ is a function that applies a feature selection alogrithm along the spectral dimension of a $p\times p\times N_{in}$ hyperspectral image volume $\mathbf{X}$ and produces another $p\times p\times N_{out}$ output volume $\mathbf{Y}$. Out of the many different possibilities for this function we have explored the identity function $I(\cdot)$ and $1\times 1$ spatial convolution in this paper. Let $\mathbf{X} = [X^{(i)}]_{i=1}^{N_{in}}$ and $\mathbf{Y} = [Y^{(j)}]_{j=1}^{N_{out}}$, i.e. let $X^i$ and $Y^j$ be the input and output channels along the spectral dimension. If $\Phi(\cdot)$ be the identity function, then $\mathbf{Y} = \Phi(\mathbf{X}) = I(\mathbf{X}) = \mathbf{X}$. If $\Phi(\cdot)$ is implemented using $1\times 1$ spatial convolution then it effectively performs the following operation.
\begin{equation}
Y^j = \sum_{i=1}^{N_{in}} w_{ji} X^i 
\end{equation}
$\forall j = 1, 2, \dots, N^{out}$. $n_b$ is a hyperparameter that can be tuned to improve performance on the validation set. The set of bands $\{B_1, B_2, \dots, B_{n_b}\}$ are passed as input to Block 2.\\

\subsubsection{Block 2: Band-specific spectral-spatial feature learning}\hfill 

Block 2 applies $n_b$ parallel networks, one on each band. Table \ref{table:IndianPines_arch} explores a variety of choices for these networks. Each convolutional and fully connected layer is followed by a ReLU (Rectified Linear Unit) layer \cite{Xavier:2011} which applies the following operation element-wise on the input volume.
\begin{eqnarray}
y =& ReLU(x) \\ \nonumber
  =& max(0,x)
\end{eqnarray} 
The outputs of the parallel networks are concatenated and fed into Block 3. \\

\subsubsection{Block 3: Summarization and classification}\hfill \\

Block 3 summarizes the concatenated outputs of the band-specific networks of Block 2 by using a set of fully connected layers, each of which is followed by a ReLU layer. A $C$-way softmax layer does the final classification by calculating the conditional probabilities of the $C$ output classes, $\mathbf{p} = [p_1, p_2, \dots, p_C ]$ as:
\begin{equation}
p_i = \frac{e^{z_i}}{\sum_{i=1}^C e^{z_i}}
\end{equation}
where $\mathbf{z} = [z_1, z_2, \dots, z_C]$ is the input to the softmax layer.

\subsection{Architectures explored}
\begin{figure}
\centering
	\includegraphics[width=\linewidth]{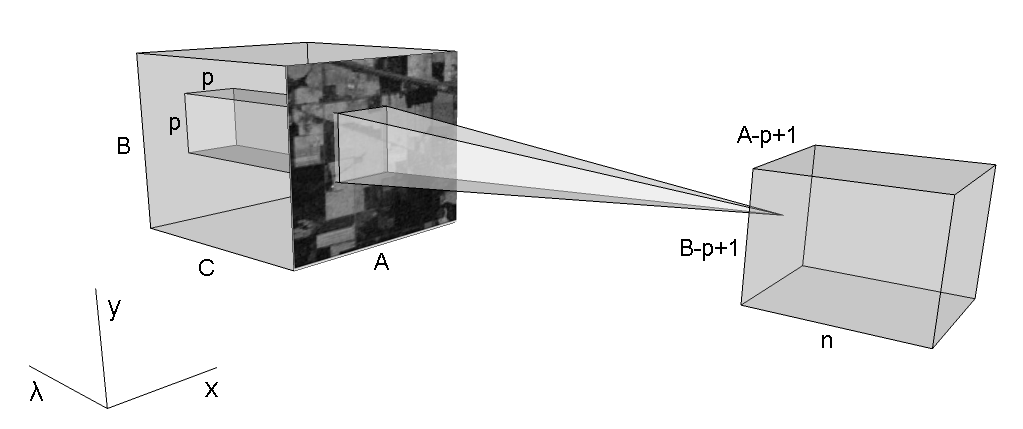}
	\caption{Diagrammatic representation of $conv_{xy}-p,n$ on a $A\times B\times C$ input volume.}
	\label{fig:spatial_conv}
\end{figure}

\begin{figure}
\centering
	\includegraphics[width=\linewidth]{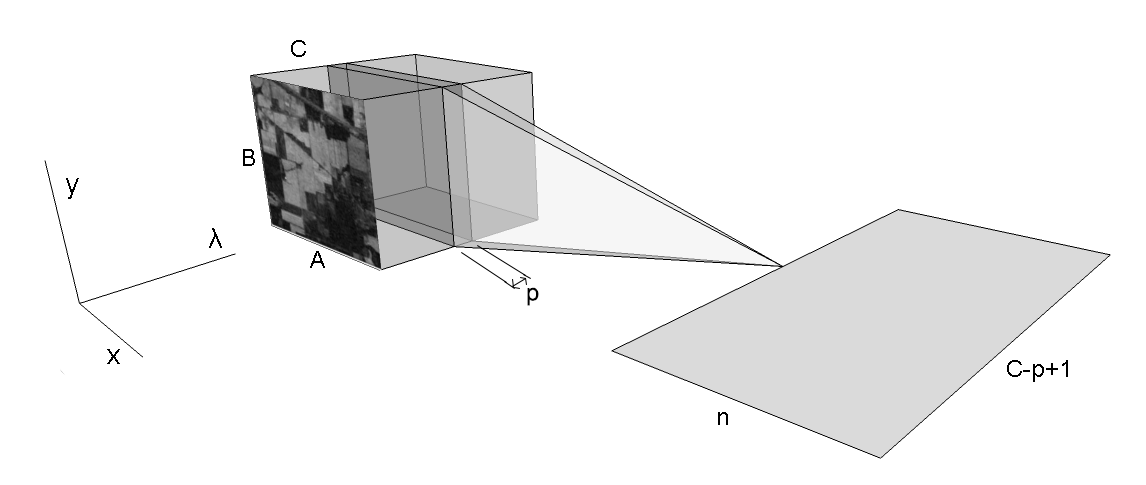}
	\caption{Diagrammatic representation of $conv_{\lambda}-p,n$ on a $A\times B\times C$ input volume.}
	\label{fig:spectral_conv}
\end{figure}

Table \ref{table:IndianPines_arch} shows four different network configurations (1-4) and their validation accuracies on the Indian Pines data set (see Section \ref{sec:data sets}) in an attempt to demonstrate the effect of different architectural design choices on the performance of the network. Only weight layers have been shown to avoid clutter. In all the four configurations the parallel networks in Block 2 have identical architecture. The Block 2 row shows the architecture of one of the parallel networks. Each $conv$ and $fc$ layer (except the last one in Block 3) is followed by a $ReLU$ layer. Cells with an asterisk (*) in the beginning mark the salient points of difference of the corresponding configuration from the one to the left of it. $PS=ON/OFF$ indicates whether parameter-sharing is on/off among the networks of Block 2. $conv_{xy}-p,n$ represents a spatial convolutional layer with receptive field size of $p\times p$ and $n$ output spectral-channels. $conv_{\lambda}-p,n$ represents a spectral convolutional layer with a spectral receptive field of size $p$ and $n$ output spatial-channels. Each convolutional layer, spatial or spectral, consists of a set of $3$-dimensional filters, one corresponding to each output channel. Each filter in a $conv_{xy}-p,n$ layer has a spatial extent of $p\times p$ and extends throughout the entire spectral axis of the input volume (Figure \ref{fig:spatial_conv}). On the other hand, a filter in a $conv_{\lambda}-p,n$ layer has a spectral extent of $p$ and extends throughout the spatial extent of the input volume (Figure \ref{fig:spectral_conv}). All convolutions used in our networks are "valid" which means there is no zero-padding of the input volume during convolution. As shown in Figures \ref{fig:spatial_conv} and \ref{fig:spectral_conv}, if we have a $A\times B\times C$ input volume then the output volumes of a $conv_{xy}-p,n$ layer and a $conv_{\lambda}-p,n$ layer with valid convolutions will respectively be $(A-p+1)\times (B-p+1)\times n$ and $n\times 1\times (C-p+1)$. $fc-n$ denotes a fully-connected layer with $n$ nodes.\\

Significance of different design choices are as follows.\\

\begin{enumerate}

\item Tying the parameters of the parallel networks in Block 2 yields an improvement of validation accuracy by at least $1\%$ in all the four configurations. This confirms that reducing the number of free parameters through parameter sharing leads to better generalization by reducing chances of overfitting.\\

\item Configuration 2 is constructed by replacing the first fully connected layer in Block 2 of Configuration 1 with a couple of spectral convolution layers. Higher validation accuracy of Configuration 2 can be attributed to fewer parameters in Block 2 than Configuration 1.\\

\item Configuration 1 and 2 have $\Phi(\cdot)=I(\cdot)$. Configuration 3 is constructed by replacing $I(\cdot)$ in Block 1 of Configuration 2 with a $1\times 1$ spatial convolution followed by $ReLU$. An improvement in validation accuracy is observed. This demonstrates the importance of a non-trivial spectral feature selection function. Such a function increases the discriminative power of the network by adding more parameters and non-linearity.\\

\item Configuration 4 is constructed by replacing the last fully connected layer in Block 2 of Configuration 3 with two spectral convolution layers and removing the first fully connected layer of Block 3. This construction improves the validation accuracy further by $1\%$ with parameter sharing in Block 2 and $0.5\%$ without. This shows that in the presence of a non-trivial spectral feature selection function in Block 1, reducing the number of parameters in Block 2 and 3 can help achieve better generalization by reducing overfitting. This also shows that adding more spectral convolution layers in Block 2 and reducing the number of fully connected layers in Block 3 leads to better performance. \\

\end{enumerate}

We use Configuration 4 with input patch-size $3\times 3$ and $PS=ON$ in all the experiments of our comparative study in Sections \ref{sec:hyperparameters} and \ref{sec:comparison} with some minor modifications for Salinas and U. Pavia data sets as listed below. 
\begin{enumerate}
\item The $1\times 1$ spatial covolution layer of Block 1 has $224$ and $100$ output channels for Salinas and U. Pavia respectively.
\item The number of parallel networks in Block 2, $n_b$, is $14$ and $5$ for Salinas and U. Pavia respectively.
\item In case of Salinas data set, the last layer of Block 3 is $fc-16$ as the number of output classes is $16$.
\end{enumerate}

\begin{table*}
\centering
\caption{comparison of different architectural design choices in terms of accuracy on the validation split of the Indian Pines data set}
\label{table:IndianPines_arch}
\begin{tabular}{c|c|c|c|c|c|c|c|c|}
\cline{2-9}
                                               & \multicolumn{2}{c|}{Configuration 1}                                                             & \multicolumn{2}{c|}{Configuration 2}                                                                  & \multicolumn{2}{c|}{Configuration 3}                                                                & \multicolumn{2}{c|}{Configuration 4}                                                                 \\ \hline
\multicolumn{9}{|c|}{Input volume: $3\times 3\times 220$}                                                                                                                                                                                                                                                                                                                                                                                                              \\ \hline
\multicolumn{1}{|c|}{Block 1}                  & \multicolumn{2}{c|}{---}                                                                         & \multicolumn{2}{c|}{---}                                                                              & \multicolumn{2}{c|}{*$conv_{xy}-1,220$}                                                             & \multicolumn{2}{c|}{$conv_{xy}-1,220$}                                                               \\ \hline
\multicolumn{9}{|c|}{Split into $n_b$ bands along the $\lambda$-axis}                                                                                                                                                                                                                                                                                                                                                                                                  \\ \hline
\multicolumn{1}{|c|}{\multirow{4}{*}{Block 2}} & $fc-150$          & \multirow{4}{*}{$n_b=10$} & *$conv_{\lambda}-3,20$ & \multirow{4}{*}{$n_b=10$} & $conv_{\lambda}-3,20$ & \multirow{4}{*}{
$n_b=10$} & $conv_{\lambda}-3,20$  & \multirow{4}{*}{$n_b=10$} \\ \cline{2-2} \cline{4-4} \cline{6-6} \cline{8-8}
\multicolumn{1}{|c|}{}                         & $fc-100$          &                                                                              & *$conv_{\lambda}-3,20$ &                                                                              & $conv_{\lambda}-3,20$ &                                                                             & *$conv_{\lambda}-3,20$ &                                                                             \\ \cline{2-2} \cline{4-4} \cline{6-6} \cline{8-8}
\multicolumn{1}{|c|}{}                         & \multirow{2}{*}{} &                                                                              & $fc-100$               &                                                                              & $fc-100$              &                                                                             & *$conv_{\lambda}-3,10$ &                                                                             \\ \cline{4-4} \cline{6-6} \cline{8-8}
\multicolumn{1}{|c|}{}                         &                   &                                                                              &                        &                                                                              &                       &                                                                             & $conv_{\lambda}-5,5$   &                                                                             \\ \hline
\multicolumn{9}{|c|}{Concatenate the outputs of the parallel networks}                                                                                                                                                                                                                                                                                                                                                                                                 \\ \hline
\multicolumn{1}{|c|}{\multirow{3}{*}{Block 3}} & \multicolumn{2}{c|}{$fc-500$}                                                                    & \multicolumn{2}{c|}{$fc-500$}                                                                         & \multicolumn{2}{c|}{$fc-500$}                                                                       & \multicolumn{2}{c|}{$fc-100$}                                                                        \\ \cline{2-9} 
\multicolumn{1}{|c|}{}                         & \multicolumn{2}{c|}{$fc-100$}                                                                    & \multicolumn{2}{c|}{$fc-100$}                                                                         & \multicolumn{2}{c|}{$fc-100$}                                                                       & \multicolumn{2}{c|}{$fc-9$}                                                                          \\ \cline{2-9} 
\multicolumn{1}{|c|}{}                         & \multicolumn{2}{c|}{$fc-9$}                                                                      & \multicolumn{2}{c|}{$fc-9$}                                                                           & \multicolumn{2}{c|}{$fc-9$}                                                                         & \multicolumn{2}{c|}{}                                                                                \\ \hline
\multicolumn{9}{|c|}{$9$-way softmax layer for classification}                                                                                                                                                                                                                                                                                                                                                                                                         \\ \hline
\multicolumn{1}{|c|}{\begin{tabular}[c]{@{}c@{}}Validation Accuracy\\ $PS=OFF$\end{tabular}}      & \multicolumn{2}{c|}{$93\%$}                                                                      & \multicolumn{2}{c|}{$95.5\%$}                                                                         & \multicolumn{2}{c|}{$97.5\%$}                                                                       & \multicolumn{2}{c|}{$98\%$}                                                                        \\ \hline
\multicolumn{1}{|c|}{\begin{tabular}[c]{@{}c@{}}Validation Accuracy\\ $PS=ON$\end{tabular}}      & \multicolumn{2}{c|}{$94\%$}                                                                      & \multicolumn{2}{c|}{$97.5\%$}                                                                         & \multicolumn{2}{c|}{$98.5\%$}                                                                       & \multicolumn{2}{c|}{$99.5\%$}                                                                        \\ \hline
\end{tabular}
\end{table*}

\subsection{Learning algorithm}
The networks are trained by minimizing the cross-entropy loss function \cite{Goodfellow:2016}. If $\mathcal{C}$ be the total number of output classes, $\{X_i,y_i\}_{i=1}^N$ be the training set, $P_{data}(class=c|X)$ and $P_{model}(class=c|X)$, $\forall c = 1, 2, \dots, \mathcal{C}$ be the observed and model conditional distributions respectively then the cross entropy loss function, $\mathcal{L}_{\times -entropy}$ is given by:

\begin{equation}
\mathcal{L}_{\times -entropy} = - \sum_{i=1}^N \sum_{c=1}^{\mathcal{C}} P_{data}(c|X_i) log (P_{model}(c|X_i))
\end{equation}

In our data sets, the observed conditional distribution $P_{data}$ is a one-hot distribution i.e.

\begin{equation}
P_{data}(class=i|X) = 
\begin{cases}
1, \quad \text{if } y=i\\
0, \quad \text{otherwise}
\end{cases}
\end{equation}

Hence the expression of cross-entropy loss function becomes:

\begin{equation}
\mathcal{L}_{\times -entropy} = - \sum_{i=1}^N log(P_{model}(y_i|X_i))
\end{equation}

Thus minimizing this expression is equivalent to maximizing the log-likelihood of the target labels given the inputs.\\

The Adam optimizer \cite{Kingma:2015} is used for making the parameter updates. It computes adaptive learning rates for each parameter. The base learning rate is set to $0.0005$ and batch-size to $100$. Dropout, with probability $0.5$ is applied to the fully connected layers of Block 3. Dropout is an effective method of regularizing neural networks by preventing co-adaptation of features \cite{Srivastava:2014}. Batchnorm \cite{Ioffe:2015} is observed to degrade the performance of our network and hence is not used.

\section{Experimental Results}
\label{sec:results}
We first present the details of the data set used; followed by the classification performances.

\subsection{data sets}
\label{sec:data sets}
The experiments are performed on three popular hyperspectral image classification data sets\footnote{http://www.ehu.eus/ccwintco/index.php?title=Hyperspectral\_ Remote\_Sensing\_Scenes} viz. Indian Pines, Salinas, and Pavia University scene (U. Pavia). Some classes in the Indian Pines data set have very few samples. We reject those classes and select the top $9$ classes by population for experimentation. The problem of insufficient samples is less severe for Salinas and U. Pavia and all the classes are taken into account. $200$ labeled pixels from each class are randomly picked to construct a training set. The rest of the labelled samples constitute the test set. A validation set is extracted from the available training set for tuning the hyperparameters of the model. As different frequency channels have different dynamic ranges, their values are normalized to the range $[0,1]$ using the following formula. 
\begin{equation}
y = \frac{x - min(x)}{max(x)}
\end{equation}

\begin{table}
\caption{data sets used}
\label{table:data sets}
\centering
\begin{tabular}{l|p{1.5cm}|p{1.5cm}|p{1.5cm}}

     & Indian Pines & Salinas & U. Pavia  \\ \hline\hline 
Sensor & AVIRIS     & AVIRIS  & ROSIS \\ \hline 
Place  & Northwestern Indiana & Salinas Valley California & Pavia, Northern Italy \\ \hline 
Frequency Band & $0.4$-$0.45 \mu m$ & $0.4$-$0.45 \mu m$ & $0.43$-$0.86 \mu m$ \\ \hline 
Spatial Resolution & $20m$ & $20m$ & $1.3m$ \\ \hline 
No. of Channels & 220 & 224 & 103 \\ \hline 
No. of Classes & 16 & 16 & 9 \\ \hline
\end{tabular}
\end{table}

\subsection{Evaluation metrics}
We evaluate the proposed architecture in terms of the following metrics. \\

\subsubsection{Class-specific accuracy} 
Class specific accuracy for class $C_i$ is calculated as the fraction of samples from class $C_i$ which were correctly classified. \\

\subsubsection{Overall accuracy} 
Overall accuracy is the ratio of the total number of correctly classified samples to the total number of samples of all classes.\\

\subsubsection{Macro and micro-averaged precision, recall and F-score}
Let $TP$, $TN$, $FN$ and $FP$ denote respectively, the number of true positive, true negative, false negative and false positive samples. Then,
\begin{equation}
Precision = \frac{TP}{TP+FP}
\end{equation}
\begin{equation}
Recall = \frac{TP}{TP+FN}
\end{equation}
\begin{equation}
F-score = \frac{2TP}{2TP+FP+FN}
\end{equation}
\indent Let $M(TP,FP,TN,FN)$ be an evaluation metric, e.g. Precision, Recall, F-score. The macro and micro averaged values of the metric can be calculated as:
\begin{equation}
M_{micro} = M(\sum_{c=1}^N TP_c, \sum_{c=1}^N FP_c, \sum_{c=1}^N TN_c, \sum_{c=1}^N FN_c)
\end{equation}
\begin{equation}
M_{macro} = \frac{1}{N}\sum_{c=1}^N M(TP_c, FP_c, TN_c, FN_c)
\end{equation}
where $N$ is the total number of output classes. A significantly lower value of the micro-average of a metric than the macro-average indicates that the less populated labels are correctly classified while the most populated labels have been grossly misclassified and vice versa \cite{Sokolova:2009}.\\

\subsubsection{$\kappa$ score}
The $\kappa$-score or $\kappa$-coefficient is a statistical measure of the degree of agreement among different evaluators \cite{Cohen:1960}. Suppose there are two evaluators that classify $N$ items into $C$ mutually exclusive classes. Then the $\kappa$-score is given by the following equation.

\begin{equation}
\kappa = \frac{p_0-p_e}{1-p_e}
\end{equation}

\noindent where $p_0$ is the relative observed probability of agreement and $p_e$ is the hypothetical probability of chance agreement. $\kappa=1$ indicates complete agreement between the evaluators while $\kappa \leq 0$ means there is no agreement at all.

\subsection{Implementation platform}
The networks are implemented in Torch\footnote{http://torch.ch}, a popular deep learning library written in Lua. The models are trained on a NVIDIA Tesla K20c GPU.

\subsection{Comparison of different hyperparameter settings}
\label{sec:hyperparameters}
\begin{figure}
\centering
	\includegraphics[width=\linewidth]{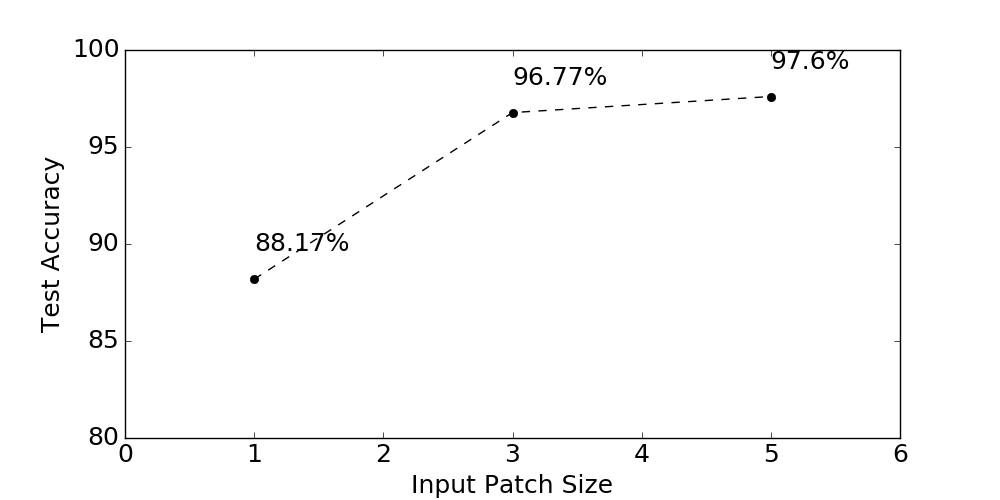}
	\caption{Variation of test accuracy on the Indian Pines data-set with input patch-size in Configuration 4.}
	\label{fig:patchsize_variation}
\end{figure}

\begin{figure}
\centering
	\includegraphics[width=\linewidth]{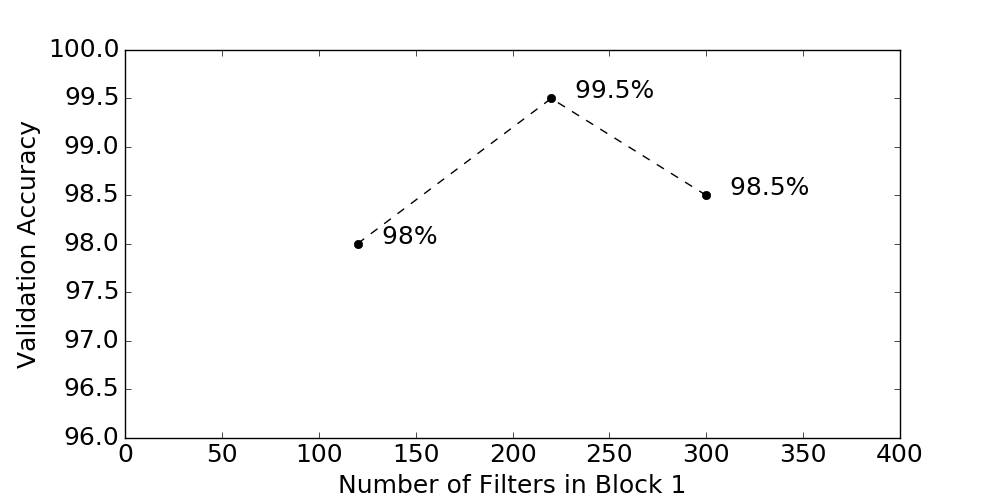}
	\caption{Variation of validation accuracy on the Indian Pines data-set with the number of output channels in Block 1 in Configuration 4.}
	\label{fig:block1_channels_variation}
\end{figure}

\begin{figure}
\centering
	\includegraphics[width=\linewidth]{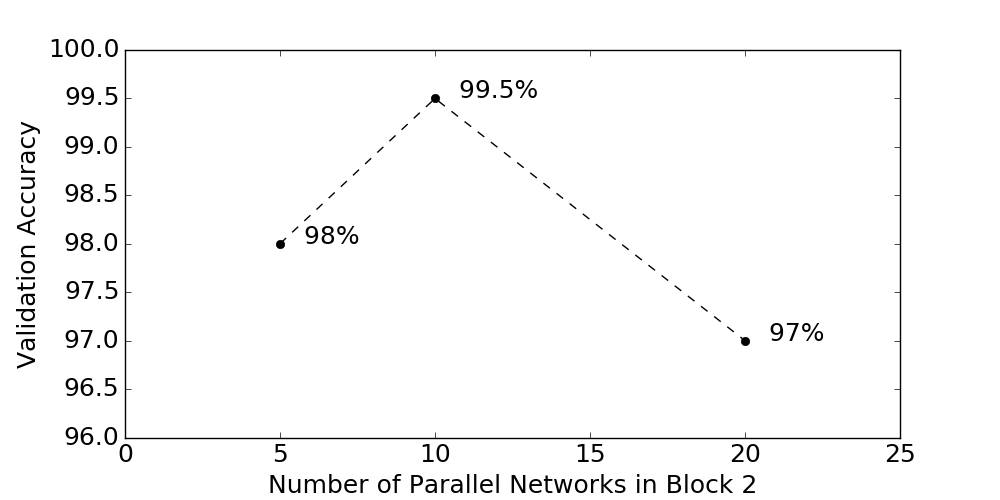}
	\caption{Variation of validation accuracy on the Indian Pines data-set with the number of parallel networks in Block 2 in Configuration 4.}
	\label{fig:block2_networks_variation}
\end{figure}

Figures \ref{fig:block1_channels_variation} and \ref{fig:block2_networks_variation} show the effect of changing the number of output channels of $1\times 1$ spatial convolution in Block 1 and the number of networks in Block 2 of Configuration 4 on validation accuracy on Indian Pines. Figure \ref{fig:patchsize_variation} shows test accuracies on Indian Pines for different choices of input patch-size. Increasing the patch-size gives more spatial context which results in marginally better accuracy of classification. However due to an increased number of parameters the model might tend to learn the data set-bias and fail to generalize to samples outside the image region from which the training and testing samples were extracted.\\

\subsection{Comparison with other methods}
\label{sec:comparison}
The test accuracies of the BASS Net architecture (BASS) on Indian Pines, Salinas and U. Pavia data sets are compared with other traditional and deep learning based classifiers in Tables \ref{table:IndianPines_acc_comp}, \ref{table:Salinas_acc_comp} and \ref{table:Pavia_acc_comp}. Among traditional classifiers $k$-Nearest Neighbor ($k$-NN), Support Vector Machine (SVM) with Random Feature Selection \cite{Waske:2010} and Extreme Learning Machine (ELM) \cite{Samat:2014,Li:2015} are compared. $k$-NN is implemented in \textit{scikit learn}\footnote{http://scikit-learn.org} while SVM is implemented using    \textit{libsvm}\footnote{http://www.csie.ntu.edu.tw/~cjlin/libsvm/}. An implementation of ELM is downloaded from the Web Page\footnote{http://www.ntu.edu.sg/home/egbhuang/elm_codes.html}. Among deep learning based classifiers, a $N_c$-$150$-$100$-$50$-$C$ Multi-Layer Perceptron (MLP), the Convolutional Neural Network (CNN) architecture of Hu et al. \cite{Hu:2015} and the Convolutional Neural Network with Pixel-Pair Features (PPF) of Li et al. \cite{Li:2016} are implemented in Torch for comparison. \\

\begin{table}
\centering
\caption{Class-specific accuracy (\%) and Overall Accuracy (OA) of different techniques for the Indian Pines data set}
\label{table:IndianPines_acc_comp}
\begin{tabular}{|c|c|c|c|c|c|c|c|}
\hline
Class & k-NN  & SVM   & ELM   & MLP   & CNN   & PPF   & BASS             \\ \hline
1     & 61.83 & 88.73 & 86.06 & 77.77 & 78.58 & 92.99 & 96.09            \\
2     & 72.65 & 91.20 & 88.19 & 79.05 & 85.23 & 96.66 & 98.25            \\
3     & 95.65 & 97.52 & 96.07 & 94.70 & 95.75 & 98.58 & 100              \\
4     & 98.90 & 99.86 & 99.73 & 98.11 & 99.81 & 100   & 99.24            \\
5     & 100   & 100   & 100   & 99.64 & 99.64 & 100   & 100              \\
6     & 80.76 & 91.67 & 90.02 & 83.68 & 89.63 & 96.24 & 94.82            \\
7     & 59.39 & 78.79 & 71.00 & 79.60 & 81.55 & 87.80 & 94.41            \\
8     & 75.72 & 93.76 & 95.62 & 89.31 & 95.42 & 98.98 & 97.46            \\
9     & 94.86 & 98.74 & 98.66 & 98.12 & 98.59 & 99.81 & 99.90            \\ \hline
OA    & 76.24 & 89.83 & 87.33 & 85.48 & 86.44 & 94.34 & \textbf{96.77} \\ \hline
\end{tabular}
\end{table}

\begin{table}
\centering
\caption{Class-specific accuracy (\%) and Overall Accuracy (OA) of different techniques for the Salinas data set}
\label{table:Salinas_acc_comp}
\begin{tabular}{|c|c|c|c|c|c|c|c|}
\hline
Class & k-NN  & SVM   & ELM   & MLP   & CNN   & PPF   & BASS             \\ \hline
1     & 98.71 & 99.55 & 99.75 & 99.67 & 97.34 & 100   & 100              \\
2     & 99.65 & 99.92 & 99.87 & 99.77 & 99.29 & 99.88 & 99.97            \\
3     & 99.09 & 99.44 & 99.60 & 98.37 & 96.51 & 99.60 & 100              \\
4     & 99.78 & 99.86 & 99.64 & 99.75 & 99.66 & 99.49 & 99.66            \\
5     & 95.28 & 98.02 & 98.81 & 98.83 & 96.97 & 98.34 & 99.59            \\
6     & 99.49 & 99.70 & 99.67 & 99.68 & 99.60 & 99.97 & 100              \\
7     & 99.55 & 99.69 & 99.66 & 99.29 & 99.49 & 100   & 99.91            \\
8     & 63.53 & 84.85 & 84.04 & 75.96 & 72.25 & 88.68 & 90.11            \\
9     & 95.94 & 99.58 & 99.89 & 99.27 & 97.53 & 98.33 & 99.73            \\
10    & 91.98 & 96.49 & 95.03 & 96.07 & 91.29 & 98.60 & 97.46            \\
11    & 98.41 & 98.78 & 96.82 & 97.93 & 97.58 & 99.54 & 99.08            \\
12    & 99.84 & 100   & 100   & 100   & 100   & 100   & 100              \\
13    & 98.69 & 99.13 & 98.25 & 99.58 & 99.02 & 99.44 & 99.44            \\
14    & 97.38 & 98.97 & 97.94 & 98.96 & 95.05 & 98.96 & 100              \\
15    & 65.66 & 76.38 & 72.96 & 75.93 & 76.83 & 83.53 & 83.94            \\
16    & 99.00 & 99.56 & 99.06 & 98.51 & 98.94 & 99.31 & 99.38            \\ \hline
OA    & 86.29 & 93.15 & 92.42 & 90.78 & 89.28 & 94.80 & \textbf{95.36} \\ \hline
\end{tabular}
\end{table}

\begin{table}
\centering
\caption{Class-specific accuracy (\%) and Overall Accuracy (OA) of different techniques for the Pavia University Scene data set}
\label{table:Pavia_acc_comp}
\begin{tabular}{|c|c|c|c|c|c|c|c|}
\hline
Class & k-NN  & SVM   & ELM   & MLP   & CNN   & PPF   & BASS             \\ \hline
1     & 77.70 & 87.95 & 81.32 & 91.73 & 88.38 & 97.42 & 97.71            \\
2     & 75.30 & 91.17 & 90.91 & 94.79 & 91.27 & 95.76 & 97.93            \\
3     & 77.27 & 86.99 & 85.09 & 85.41 & 85.88 & 94.05 & 94.95            \\
4     & 92.46 & 95.50 & 96.61 & 94.13 & 97.24 & 97.52 & 97.80            \\
5     & 99.63 & 99.85 & 99.63 & 99.65 & 99.91 & 100   & 100              \\
6     & 79.50 & 94.31 & 94.33 & 90.87 & 96.41 & 99.13 & 96.60            \\
7     & 92.86 & 94.74 & 95.94 & 92.56 & 93.62 & 96.19 & 98.14            \\
8     & 76.45 & 85.89 & 82.65 & 83.19 & 87.45 & 93.62 & 95.46            \\
9     & 99.62 & 99.89 & 99.79 & 99.73 & 99.57 & 99.60 & 100              \\ \hline
OA    & 79.45 & 91.10 & 89.86 & 92.54 & 92.27 & 96.48 & \textbf{97.48} \\ \hline
\end{tabular}
\end{table}

\begin{table}
\centering
\caption{classification performance statistics}
\label{table:kappa_scores}
\begin{tabular}{cc|c|c|c|}
\cline{3-5}
                                                      &           & Indian Pines & Salinas & U. Pavia \\ \hline
\multicolumn{1}{|c|}{\multirow{3}{*}{micro-averaged}} & precision & 0.9677       & 0.9536  & 0.9748   \\ \cline{2-5} 
\multicolumn{1}{|c|}{}                                & recall    & 0.9677       & 0.9536  & 0.9748   \\ \cline{2-5} 
\multicolumn{1}{|c|}{}                                & F-score   & 0.9677       & 0.9536  & 0.9748   \\ \hline
\multicolumn{1}{|c|}{\multirow{3}{*}{macro-averaged}} & precision & 0.9713       & 0.9730  & 0.9680   \\ \cline{2-5} 
\multicolumn{1}{|c|}{}                                & recall    & 0.9779       & 0.9802  & 0.9762   \\ \cline{2-5} 
\multicolumn{1}{|c|}{}                                & F-score   & 0.9745       & 0.9764  & 0.9719   \\ \hline
\multicolumn{2}{|c|}{$\kappa$- score}                             & 0.9612       & 0.9480  & 0.9662   \\ \hline
\end{tabular}
\end{table}

\subsection{Results and discussion}
\begin{figure}
\centering
	\includegraphics[width=\linewidth]{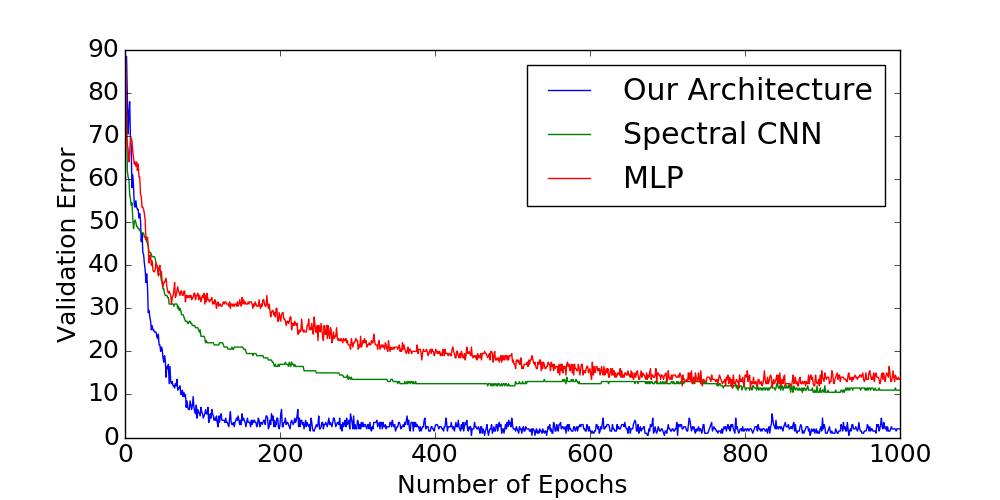}
	\caption{Variation of validation error over epochs of training on Indian Pines data set for the proposed architecture and other popular deep neural networks.}
	\label{fig:convergence}
\end{figure}

Tables \ref{table:IndianPines_acc_comp}, \ref{table:Salinas_acc_comp} and \ref{table:Pavia_acc_comp} show the results of the comparison of the proposed framework with traditional and deep learning based methods. The proposed framework outperforms all the other methods on all the three data sets in terms of Overall Accuracy (OA) of classification. For example, on Indian Pines, the test-accuracy of our network exceeds SVM, CNN and PPF by $6.94\%$, $10.33\%$ and $2.43\%$ respectively. Figure \ref{fig:convergence} compares the variation of validation accuracy over epochs of training on Indian Pines. Our network converges faster than MLP and CNN. Table \ref{table:kappa_scores} gives micro and macro-averaged precision, recall and F-score and $\kappa-$ scores for our models trained on the three data sets. High values of both macro and micro-averaged classification metrics viz. precision, recall and F-score suggest that the classifier is effective for both scarce and abundant classes. High values of $\kappa$-score for all the data sets show that the proposed classifier has high degree of agreeability with the ground truth generating mechanism.

\begin{figure}
\centering
	\includegraphics[width=\linewidth]{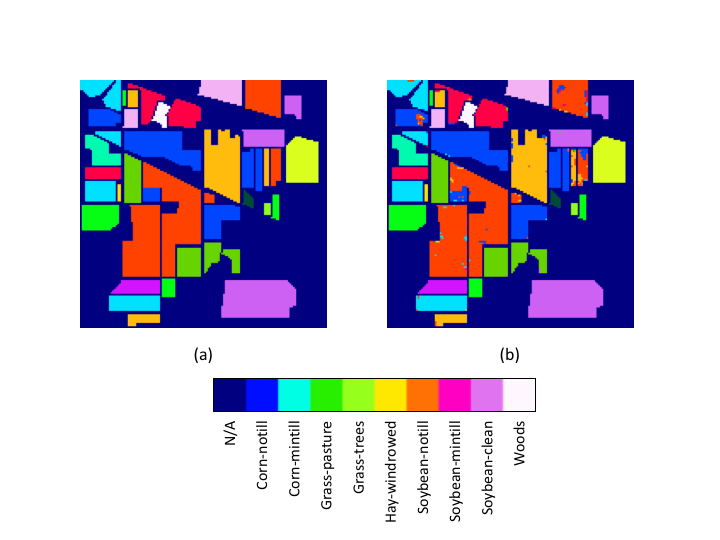}
	\caption{Thematic maps resulting from classification for the Indian Pines data set with $9$ classes. (a) ground-truth map (b) decoded output from our model.}
	\label{fig:IndianPines_decoding}
\end{figure}%

\begin{figure}
\centering
	\includegraphics[width=\linewidth]{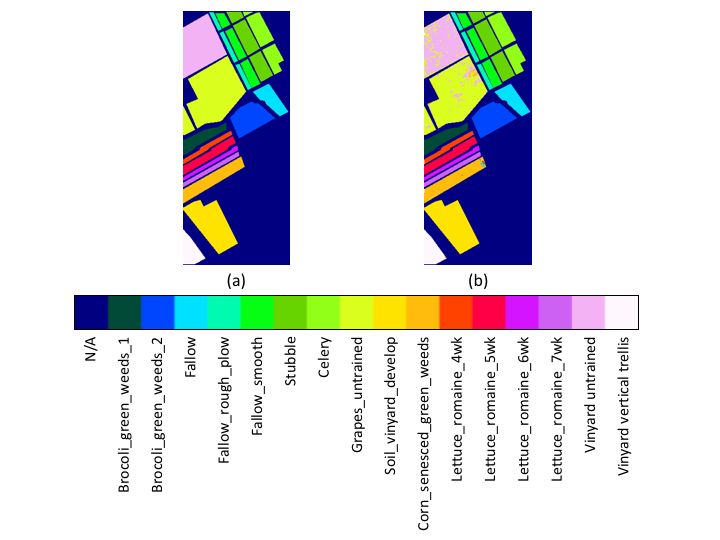}
	\caption{Thematic maps resulting from classification for the Salinas data set with $16$ classes. (a) ground-truth map (b) decoded output from our model.}
	\label{fig:Salinas_decoding}
\end{figure}%

\begin{figure}
\centering
	\includegraphics[width=\linewidth]{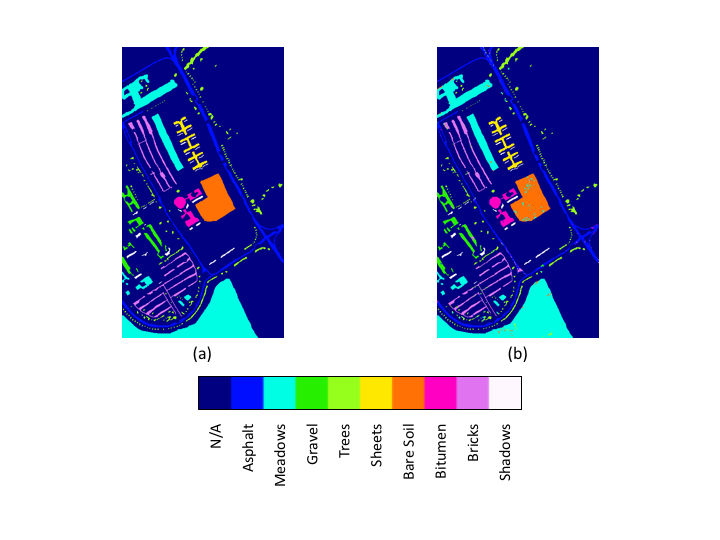}
	\caption{Thematic maps resulting from classification for the Pavia University Scene data set with $9$ classes. (a) ground-truth map (b) decoded output from our model.}
	\label{fig:Pavia_decoding}
\end{figure}%

\section{Conclusion}
\label{sec:conclusion}
In this paper an end-to-end deep learning neural network architecture has been proposed that performs band-specific spectral-spatial feature learning for superior modeling of spectral signatures. Curse of dimensionality and scarcity of labeled training examples are tackled by extensive parameter sharing in the network. Predictions are made on the basis of a $p\times p$ neighborhood around the target pixel in order to take care of large spatial variability of spectral signature in hyperspectral images. Experiments on benchmark hyperspectral image classification data sets show superior classification performance and faster convergence than other popular deep learning based methods. 

\section{Acknowledgement}
\label{sec:acknowledgement}
This study was performed as a part of the project titled "Deep Learning for Automated Feature Discovery in Hyperspectral Images (LDH)" sponsored by Space Applications Centre (SAC), Indian Space Research Organization (ISRO). Anirban Santara's work in this project was supported by Google India under the Google India PhD Fellowship Award.

\bibliography{my_citations.bib}
\bibliographystyle{ieeetr}

\end{document}